\documentclass[10pt,twocolumn,letterpaper]{article}

\usepackage{iccv}              

\definecolor{iccvblue}{rgb}{0.21,0.49,0.74}
\usepackage[pagebackref,breaklinks,colorlinks,allcolors=iccvblue]{hyperref}
\usepackage{multirow}

\def\paperID{*****} 
\def\confName{ICCV}
\def\confYear{2025}

\title{Visual Adaptive Prompting for Compositional Zero-Shot Learning}

\author{
Kyle Stein$^{1}$ \quad
Andrew Arash Mahyari$^{2}$ \quad
Guillermo Francia III$^{1}$ \quad
Eman El-Sheikh$^{1}$\\
$^{1}$University of West Florida, Pensacola, FL, USA\\
$^{2}$Florida Institute for Human and Machine Cognition (IHMC), Pensacola, FL, USA\\
{\tt\small ks209@students.uwf.edu, amahyari@ihmc.org, gfranciaiii@uwf.edu, eelsheikh@uwf.edu}
}

\begin{document}
\maketitle

\begin{abstract}
Vision-Language Models (VLMs) have demonstrated impressive multimodal capabilities in learning joint representations of visual and textual data, making them powerful tools for tasks such as Compositional Zero-Shot Learning (CZSL). CZSL requires models to generalize to novel combinations of visual primitives—such as attributes and objects—that were not explicitly encountered during training. Recent works in prompting for CZSL have focused on modifying inputs for the text encoder, often using static prompts that do not change across varying visual contexts. However, these approaches struggle to fully capture varying visual contexts, as they focus on text adaptation rather than leveraging visual features for compositional reasoning. To address this, we propose a \textit{\textbf{V}isual \textbf{A}daptive \textbf{P}rompting \textbf{S}ystem (VAPS)} that leverages a learnable visual prompt repository and similarity-based retrieval mechanism within the framework of VLMs to bridge the gap between semantic and visual features. Our method introduces a dynamic visual prompt repository mechanism that selects the most relevant attribute and object prompts based on the visual features of the image. Our proposed system includes a visual prompt adapter that encourages the model to learn a more generalizable embedding space. Experiments on three CZSL benchmarks, across both closed and open-world scenarios, demonstrate state-of-the-art results.

   
\end{abstract}

\section{Introduction}
\label{sec:intro}

Humans have a remarkable ability to compose attributes with objects to imagine novel combinations they have never encountered, for example, a blue banana. Attributes describe the state of an object, such as the color, texture, or shape, while objects represent the entities themselves, such as a banana or a car. This ability to associate various attributes with different objects is a fundamental aspect of human cognition, known as compositionality \cite{lake2014towards, lake2017building, frankland2020concepts}. Compositional Zero Shot Learning (CZSL) aims to mimic this human behavior by enabling models to recognize combinations of primitive attributes and objects that were not explicitly composed, or seen together, during training. CZSL focuses on the recombination of known primitives, allowing for the recognition of novel compositions by effectively disentangling attribute and object information from a combined visual representation \cite{hao2023learning, huynh2020fine, yang2022decomposable}.


\begin{figure}[t!]
    \centering
    \includegraphics[width=1.0\columnwidth]{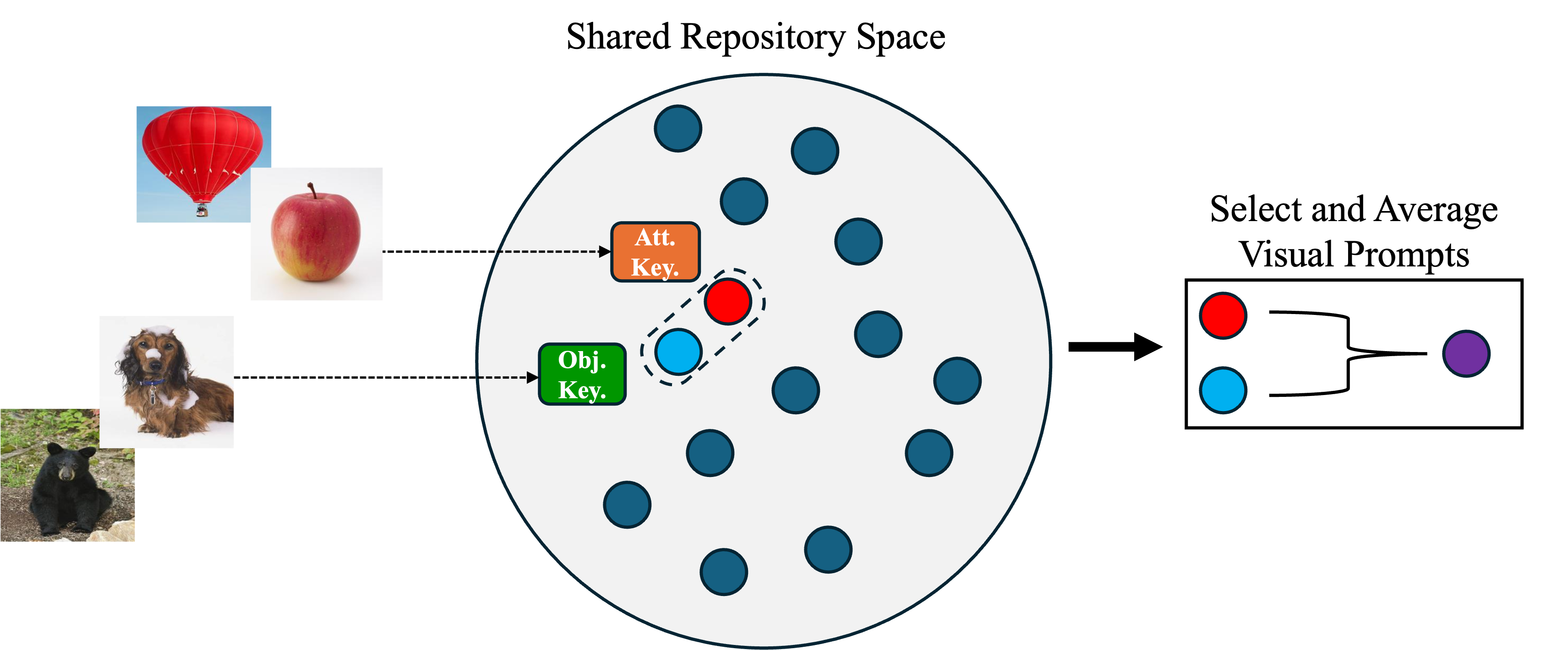} 
    \caption{Similarity-Based Visual Repository Mechanism. The image features are matched against learned keys in a shared repository space, where similarity scores are computed. Based on these similarity scores, the most relevant learned visual prompts representing attributes (i.e., “color") and objects (i.e., “animals”) are selected. The selected prompts are then averaged, creating a final representation (purple) that holds the object and attribute information of the input images. This process allows dynamic retrieval and combination of relevant visual prompts for downstream tasks.}
    \label{fig:CZSL_formation}
\end{figure}

The rigorous pre-training of VLMs have shown great ability for understanding tasks that require multimodal visual and textual data. Models like CLIP \cite{radford2021learning} are trained on vast amounts of image-text data and learn to map visual and textual representations into a shared embedding space, enabling them to perform tasks such as zero-shot classification and image recognition. However, when approached with the more nuanced understanding of compositional reasoning, CLIP lacks the ability to generalize to unseen primitive compositions. This lack of generalization can be attributed to the reliance on static representations of fixed classes seen during training and the lack of flexibility to adjust their internal prompts to changes in visual inputs \cite{jin2024llms, jin2021good, zhou2022learning}.

Recent studies in CZSL based on VLMs \cite{xu2024gipcol,bao2023prompting, lu2023decomposed} suffer from several drawbacks which limit their performance. \textbf{First}, state-of-the-art (SOTA) methods typically utilize either fixed templates ``{\sffamily{a photo of [attribute][object]}}'' or a single learned prefix prepended to ``{\sffamily{[attribute][object]}}''. In these strategies, the text prompt is processed through the text encoder, while the raw image is handled by the image encoder, leading to the development of a joint embedding that facilitates the inference of unseen attribute-object compositions. Although promising, text-focused methods largely overlook valuable visual insights, as they concentrate on tuning text-based prompts rather than exploiting visual information that could enhance attribute-object disentanglement and improve model adaptability to unseen compositions \cite{ma2018disentangled, min2020domain}. \textbf{Second}, SOTA algorithms typically rely solely on these text-centric learnable prompts, and operate under the assumption that a minimal number of prompts can adequately capture all attribute-object combinations. For example, a single prompt might be used to denote a variety of attributes like "wet," "dry," "red," etc. This limited approach constrains the development of tailored prompts that could significantly improve performance. Moreover, using only one or two prompts means these techniques struggle to properly separate attribute features from object features, thereby restricting their effectiveness in generalizing to unseen compositions. \textbf{Third}, text-centric approaches utilize fixed prompts as a prefix to ``{\sffamily{[attribute][object]}}'' during training. These \textit{\textbf{static prompts}} are comprised of a set of learnable variables that remain constant in their positions across different attribute-object pairings, such as transitioning from ``wet cat'' to ``red apple''. Relying on such static, learnable text prompts often fails to fully encompass the entire context of an image. This is because text descriptions can be relatively rigid, not adequately reflecting the complexities of an image. For instance, the attribute ``wet'' might carry different semantic implications when associated with disparate objects like a cat versus an apple.


To address these challenges, we propose a novel approach \textit{Visual Adaptive Prompting System (VAPS)} that builds on CLIP’s multi-modal architecture by leveraging a dynamic visual prompt repository and a similarity-based retrieval mechanism, which shifts the emphasis to the image features generated by CLIP’s visual encoder. VAPS creates a repository of visual features to serve as visual prompts, comparing them with fused features in a pair space, as well as implementing a prompt adapter based on the original image features that allows the model to adapt its representation based on the visual context. The summary of our  contributions for CZSL are stated below:

\begin{itemize}
    
    \item \textit{Visual Prompts:} To leverage information directly from the image encoder, we introduce visual prompts. These visual prompts are learnable embeddings designed to capture visual patterns related to attributes and objects. This high-level semantic representation efficiently separates attributes from object visual features, allowing VAPS to generalize visual semantics to unseen compositions more effectively. Unlike text-based prompts, visual prompts leverage visual features to enhance disentanglement and boost adaptability.

    \item \textit{Prompts Repository:} Our approach employs a repository of learnable visual prompts that operate independently from those used by the text encoder. Each visual prompt in this repository is paired with a learnable key, which serves as an identifier for effective selection. VAPS uses a similarity-based retrieval mechanism to match image features with learned keys in the repository, selecting the most relevant prompts for the input image. This allows the model to effectively disentangle attributes from objects, facilitating generalization to unseen compositions.

    
    \item \textit{Text Prompt Adapter:} VAPS incorporates an adapter that dynamically updates the prefix of the text prompt using image features from the visual transformer. By incorporating a bias term from image features, this approach customizes the prompt for each image, aligning it with the visual context and effectively separating attributes from objects. For instance, the learnable text prompts are adjusted with different bias values when processing images of a ``wet cat'' versus a ``red apple,'' overcoming the limitations associated with \textit{\textbf{static prompts}}.
    
\end{itemize}

\vspace{-2.5mm}
\section{Related Work}
\label{sec:related_work}
\vspace{-2mm}

\begin{figure*}[t]
    \centering
    \includegraphics[width=\textwidth, height=0.25\textheight, keepaspectratio]{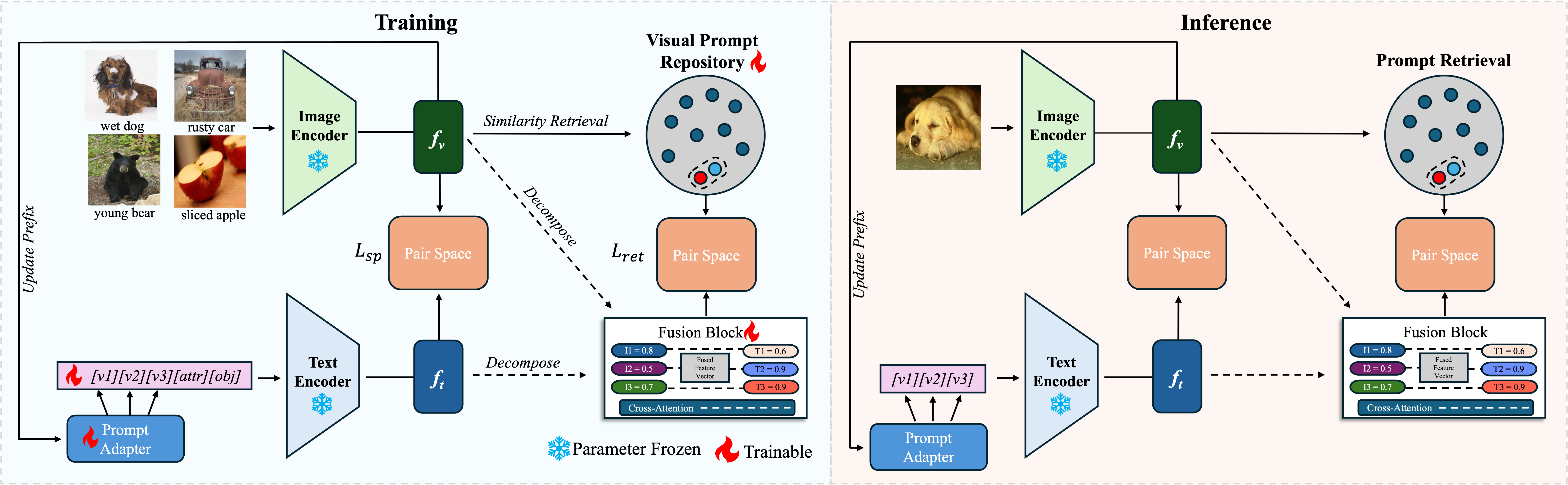} 
    \caption{VAPS leverages the multimodal power of CLIP’s image and text encoders to extract visual features $f_v\in\mathbb{R}^d$ and text features $f_t\in\mathbb{R}^{768}$. A prompt adapter uses $f_v$ to shift three learnable prefix tokens [v1][v2][v3], which are prepended to the embeddings of all attribute–object candidates and encoded to yield $f_t$. The top two visual prompts most similar to $f_v$ are retrieved and averaged, and $f_v$ and $f_t$ are projected into a shared pair space, fused with the averaged prompts, and scored to produce final logits for selecting the highest-scoring attribute–object pair.}
    \label{fig:MainFig}
\end{figure*}

\noindent\textbf{Compositional Zero-Shot Learning} extends the principles of zero-shot learning by focusing on the recognition of unseen compositions of known primitives. As previously mentioned, disentanglement is a prevalent approach in many CZSL methods \cite{Tong_2019_CVPR, chen2021semantics, li2021generalized, hao2023learning}. However, this is not the only approach to achieve compositional generalization. Li et. al. uses the principles of symmetry and group theory to model attribute-object compositions through coupling and decoupling transformations, and introduces a novel distance method for CZSL \cite{li2020symmetry}. A Siamese Contrastive Embedding Network (SCEN) embeds visual features into a Siamese contrastive space to separately capture attribute and object prototypes diversity \cite{li2022siamese}. A retrieval-augmented approach was proposed to enhance the recognition of unseen primitive compositions by retrieving and augmenting attribute and object representations \cite{jing2024retrieval}. Wang et. al. propose a dependent approach for CZSL that generates conditional attribute embeddings by using an attribute hyper learner and base learner to account for the varying interaction of attributes with different objects \cite{wang2023learning}.

Modern applications in CZSL include adapting pre-trained multimodal VLMs, such as CLIP \cite{radford2021learning}, to improve CZSL results. It is shown that downstream tasks can be built on top of the VLMs to enhance these results. Compositional Soft Prompting (CSP), introduced in \cite{nayak2022learning}, uses a static prompt prefix combined with learned attribute and object descriptions. This text is passed through a text encoder while the image is processed by CLIP’s visual encoder. The model then calculates the cosine similarity between the text embeddings and image features to predict the correct attribute-object composition. More recent works built on top of this method by removing the static prefix content and instead making the entire prompt learnable \cite{xu2024gipcol, lu2023decomposed}. While these processes produce promising results, only one learned prompt may not generalize well to every image passed through the visual encoder. 

\vspace{0mm}
\noindent\textbf{Prompt Learning/Tuning} modifies the original input by leveraging learnable tokens that guide the pre-trained language model to examine specific features or contexts relevant to the task the model is trying to solve \cite{liu2023pre, khattak2023maple, shi2023logoprompt}. With the more recent advancements in VLMs, prompt learning has steered into a new direction by focusing on the multi-modality of both textual and visual content in a shared embedding space \cite{radford2021learning, ramesh2021zero, kim2021vilt}. Huang et. al. introduced a method to transfer performance from VLMs without the need for prompt engineering or labeled data by generating pseudo labels for target datasets and optimizing learnable prompt representations through self-training on the psuedo-labeled samples \cite{huang2022unsupervised}. Prompt learning has been applied on top of pre-trained vision transformers to solve the catastrophic forgetting problem in continual learning by using a pool of learnable prompts to learn tasks sequentially \cite{wang2022learning}. CoOp \cite{zhou2022learning} introduced a method to automate prompt engineering for models like CLIP by learning the vectors of prefix content while keeping the pre-trained model fixed for few-shot scenarios. CoCoOp \cite{zhou2022conditional} learns a lightweight neural network that generates dynamic prompts based on the input image. 

\vspace{-2mm}
\section{Preliminaries and Insights}
\vspace{-2mm}
Let $A$ denote the set of attributes and $O$ denote the set of objects. Given a set of attributes $A=\{a_0, a_1, \ldots, a_n\}$ and a set of objects $O=\{o_0, o_1, \ldots, o_m\}$, we define the set of all possible compositions $C=A\times O$, where $\times$ is the Cartesian product. The set $C$ can be divided into two disjoint subsets: seen compositions $C_s$ and unseen compositions $C_u$, where $C_s \cap C_u = \emptyset$ and $C_s \cup C_u = C$.

In Closed-World CZSL, the model operates under the assumption that all possible compositions for testing are drawn from a predefined subset $C_{\text{test}} \subseteq C$. This means that while some compositions may be unseen during training, they are still part of this known subset. Consequently, the test set includes both $C_s$ and $C_u$, but all testing samples are limited to this established set $C_s \cup C_u$, restricting the model to a predefined range of feasible combinations. 
In Open-World CZSL \cite{mancini2021open}, the model must navigate the entire composition space $C = A \times O$, which includes both feasible and infeasible combinations. This presents an added challenge, as the model encounters compositions that were neither seen during training nor predefined as possible during testing, making classification more complex. The objective is to learn a function $f: X \to C_{\text{test}}$ for Closed-World settings and ${f: X \to C}$ for Open-World settings, where $X$ denotes the input space of images corresponding to these compositions.

\vspace{-2mm}
\section{Methodology}
\label{sec:methodology}
Our approach builds upon the multimodal capabilities of VLMs, such as CLIP, for compositional learning by freezing its pre-trained visual and text encoders. During training, each input image is processed through CLIP's visual encoder to obtain feature representations $f_v$. These features are used two-fold: (i) as input to the text prompt adapter to find the appropriate bias for dynamically shifting a set of learnable prefix text tokens and (ii) used to select visual prompts from the repository. The shifted prefix tokens are prepended to the word embeddings of attributes and objects to form the final text prompt, which is processed through the text encoder to produce text features $f_t$. Next, $f_v$ and $f_t$ are decomposed and fused via cross-attention, resulting in a joint representation that is mapped to a dedicated pair space. In this space, the similarity between the fused representation of $<f_v,f_t>$and the selected visual prompts informs the final compositional prediction. Additionally, both $f_v$ and $f_t$ are each projected into a separate pair space to facilitate the compositional prediction.

During inference, the same processing pipeline is followed with one main difference: ground truth attribute-object pairings are unavailable. Instead, text prompts are generated for each candidate pair \((a, o) \in \mathcal{C}^{\text{test}}\) using the shifted prefix, and the final predicted composition is selected as the pair yielding the highest probability \(p_{\text{sp}}\). Figure \ref{fig:MainFig} provides an overview of our method.

\vspace{-1mm}
\subsection{Visual Prompt Repository}
\vspace{-1mm}
The visual prompt repository comprises a collection of $M$ learnable visual prompts $\mathbf{P}_1, \mathbf{P}_2, \dots, \mathbf{P}_{M}$, where each $P_i \in \mathbb{R}^{l \times d}$ represents an individual visual prompt, and $l$ denotes the prompt length. These prompts are initialized randomly and refined during training to capture high-level semantic visual features, such as colors, textures, and shapes. Each visual prompt $P_i$ is paired with a learnable key $a_i \in \mathbb{R}^d$, which helps identify the most relevant prompts for a given image by measuring similarity between the input image features and the keys. The keys $a_i$ are used for similarity assessment, while $P_i$ contributes to predicting attributes and objects in the joint embedding space, combined with the output of the fusion block. To find the best-matching visual prompts, cosine similarity is computed between the normalized visual features $f_v$ of the input image and each normalized key $a_i$. Based on the similarity scores, the model selects the top two prompts with the highest cosine similarity, ensuring that one prompt aligns with the image’s attribute and the other with the object. These selected prompts are then averaged, forming a combined representation that is later integrated with the visual and textual features to improve compositional prediction. By dynamically selecting the most relevant visual prompts, the model improves image-text alignment and enriches the representation of image content. Over the course of training, each visual prompt becomes more adept at capturing the visual characteristics of basic elements, such as “red” or “wet,” enabling more precise attribute-object mapping and enhancing generalization to unseen compositions. The final representation of the retrieved visual prompts can be denoted by $f_{ret}$.

\subsection{Text Prompt Adapter}


In SOTA CZSL algorithms, ``{\sffamily{[attribute][object]}}'' is prepended with learnable text soft prompts. These soft prompts typically consist of a few trainable tokens, such as three prefix tokens [v1][v2][v3], which are initialized with a generic phrase like ``a photo of'' to align with CLIP's pretraining \cite{nayak2022learning,lu2023decomposed, xu2024gipcol}. However, these tokens remain fixed in location and combination for every training and inference sample, making no distinction between different attributes and objects. Therefore, during inference, when ``{\sffamily{[attribute][object]}}'' is not available, the same trained prefix text soft prompt is used for every test sample, leading to poor generalization on unseen compositions. 

Motivated by the prompting method introduced in \cite{zhou2022conditional}, we incorporate a prompt adapter module with trainable parameters. The prompt adapter takes the image feature $f_v$ as the input and provides the amount of the shift for the text prompt in the output. The prefix structure of the learnable soft prompt provides a general context for the task while the attribute and object represent the composition of the object. The prompt adapter is a lightweight neural network, represented as:

\vspace{-5mm}
\begin{equation}
\text{PromptNet}(f_v) = {\bf W}_2 \cdot \sigma({\bf W}_1 \cdot f_v + {\bf b}_1) + {\bf b}_2 ,
\end{equation}

\noindent where $f_v$ represents the visual features, ${\bf W}_1$, ${\bf W}_2$, ${\bf b}_1$, and ${\bf b}_2$ represent the linear layers and their bias terms, respectively, and $\sigma(\cdot)$ is the ReLU activation function. The output of the prompt adapter network is a bias term, denoted as $\varphi(f_v)$, which is added to each of the learnable embeddings in the soft prompt's prefix $\{\theta_0, \theta_1, \dots, \theta_p\}$. This is represented as:

\vspace{-2mm}

\begin{equation}
\theta'_i = \theta_i + \varphi_i(f_v) \quad \text{for} \quad i = 0, \dots, p,
\end{equation}


\noindent where each \( \theta'_i \) represents the shifted version of the original prompt embedding \( \theta_i \). Therefore, the updated shift in the soft prompt \( P'_{\text{soft}} \) is $P'_{\text{soft}} = \{ \theta'_0, \theta'_1, \dots, \theta'_p, \theta_a, \theta_o \}$. The text features $f_{t}$ are obtained by passing $P'_{\text{soft}}$ through the text encoder.

\subsection{Decomposition and Fusion Block}

To disentangle the visual features of attributes and objects and embed them jointly with their text representation, we decompose and fuse the visual features, $f_v$, and the text features, $f_t$ \cite{lu2023decomposed}. Specifically, the text feature representation is decomposed by averaging the contributions of the attributes and objects from the corresponding logits. Decomposition helps isolate the properties of attributes and objects, allowing the model to treat these two components independently during subsequent fusion. The decomposed features are supervised during training to accurately capture the primitive's information. We compute the attribute and object probability as follows:

\begin{equation}
p(y = a \mid x; \theta) = \frac{\exp(f_v \cdot f_{t})}{\sum\limits_{\bar{a} \in \mathcal{A}} \exp(f_v \cdot f_{t})},
\end{equation}

\begin{equation}
p(y = o \mid x; \theta) = \frac{\exp(f_v \cdot f_{t})}{\sum\limits_{\bar{o} \in \mathcal{O}} \exp(f_v \cdot f_{t})},
\end{equation}

\noindent where \( \mathcal{A} \) and \( \mathcal{O} \) denote the sets of attributes and objects. The cross-entropy can be minimized as:

\vspace{-3mm}
\begin{equation}
\mathcal{L}_{\text{att}} = - \frac{1}{| \mathcal{A}|} \sum_{(x, y) \in {C}^s} \log \left( p_{}\left( y = (a) \middle| x; \theta \right) \right),
\end{equation}

\vspace{-2mm}
\begin{equation}
\mathcal{L}_{\text{obj}} = - \frac{1}{| \mathcal{O}|} \sum_{(x, y) \in {C}^s} \log \left( p_{}\left( y = (o) \middle| x; \theta \right) \right).
\end{equation}

Next, $f_v$ and $f_{t}$ are fused through a cross-attention mechanism, where the query ($Q$), key ($K$), and value ($V$) matrices focus on aligning image and text features within a compositional space. Specifically, $Q$ is derived from $f_t$, while $K$ and $V$ are derived from $f_v$ where the key represents the aspects of the image that the query will attend to, and the value holds the information that will be emphasized based on how well $Q$ aligns with $K$. Here, $d$ denotes the dimensionality of the feature vectors in the compositional space. The cross-attention can be computed as follows:

\vspace{-3mm}
\begin{equation}
\text{Attention}(Q, K, V) = \text{softmax} \left( \frac{Q K^{T}}{\sqrt{d}} \right) V.
\end{equation}

This cross-attention operation yields $f_{t \rightarrow v}$, a fused representation that incorporates textual context of the attributes and objects with the visual features. 

\subsection{Training}

\begin{table*}[t]
\centering
\begin{tabular}{l|cccc|cccc|cccc}
\toprule
\multirow{2}{*}{Method} & \multicolumn{4}{c|}{MIT-States} & \multicolumn{4}{c|}{UT-Zappos} & \multicolumn{4}{c}{C-GQA} \\
\cmidrule{2-13}
& S & U & H & AUC & S & U & H & AUC & S & U & H & AUC \\
\midrule
AoP \cite{nagarajan2018attributes} & 14.3 & 17.4 & 9.9 & 1.6 & 59.8 & 54.2 & 40.8 & 25.9 & 17.0 & 5.6 & 5.9 & 0.7 \\
LE+ \cite{naeem2021learning}      & 15.0 & 20.1 & 10.7 & 2.0 & 53.0 & 61.9 & 41.0 & 25.7 & 18.1 & 5.6 & 6.1 & 0.8 \\
TMN \cite{purushwalkam2019task}    & 20.2 & 20.1 & 13.0 & 2.9 & 58.7 & 60.0 & 45.0 & 29.3 & 23.1 & 6.5 & 7.5 & 1.1 \\
SymNet \cite{li2020symmetry}    & 24.2 & 25.2 & 16.1 & 3.0 & 49.8 & 57.4 & 40.4 & 23.4 & 26.8 & 10.3 & 11.0 & 2.1 \\
CompCos \cite{mancini2021open}   & 25.3 & 24.6 & 16.4 & 4.5 & 59.8 & 62.5 & 43.1 & 28.1 & 28.1 & 11.2 & 12.4 & 2.6 \\
CGE \cite{naeem2021learning} & 28.7 & 25.3 & 17.2 & 5.1 & 56.8 & 63.6 & 41.2 & 26.4 & 28.7 & 25.3 & 17.2 & 5.1 \\
Co-CGE \cite{mancini2022learning}    & 32.1 & 28.3 & 20.0 & 6.6 & 62.3 & 66.3 & 48.1 & 33.9 & 33.3 & 14.9 & 14.4 & 4.1 \\
SCEN \cite{li2022siamese}      & 29.9 & 25.2 & 18.4 & 5.3 & 63.5 & 63.1 & 47.8 & 32.0 & 28.9 & 25.4 & 17.5 & 5.5 \\
\midrule
CLIP \cite{radford2021learning}  & 30.2 & 40.0 & 26.1 & 11.0 & 15.8 & 49.1 & 15.6 & 5.0 & 7.5 & 25.0 & 8.6 & 1.4 \\
CSP \cite{nayak2022learning}       & 46.6 & 49.9 & 36.3 & 19.4 & 64.2 & 66.2 & 46.6 & 33.0 & 28.8 & 26.8 & 20.5 & 6.2 \\
GIPCOL \cite{xu2024gipcol}    & \textbf{48.5} & 49.6 & 36.6 & 19.9 & \textcolor{blue}{65.0} & 68.5 & \textcolor{blue}{48.8} & \textcolor{blue}{36.2} & 31.9 & 28.4 & 22.5 & 7.1 \\
DFSP \cite{lu2023decomposed} & 46.9 & \textcolor{blue}{52.0} & \textcolor{blue}{37.3} & \textcolor{blue}{20.6} & \textbf{66.7} & \textcolor{blue}{71.7} & 47.2 & 36.0 & \textcolor{blue}{38.2} & \textbf{32.0} & \textcolor{blue}{27.1} & \textcolor{blue}{10.5} \\
\midrule
VAPS (Ours) & \textcolor{blue}{48.4} & \textbf{52.2} & \textbf{38.2} & \textbf{21.2} & 64.5 & \textbf{74.3} & \textbf{53.8} & \textbf{40.1} & \textbf{39.6} & \textcolor{blue}{31.7} & \textbf{28.1} & \textbf{11.0} \\
\bottomrule
\end{tabular}
\caption{Closed-World Results on MIT-States, UT-Zappos, and CGQA. The results are reported for Seen (S), Unseen (U), Harmonic Mean (H), and Area Under the Curve (AUC). Bold and blue indicate the first and second best results, respectively.}
\label{fig:closed}
\end{table*}

The additional training of our model is conducted in two parts: one focuses on adapting the soft prompt to the target compositions, and the other on the alignment between retrieved prompts and the final fused representation $f_{t \rightarrow v}$. To ensure the shifted soft prompts align with the target compositions, the class probability for the soft prompt are computed as follows:

\vspace{-3.5mm}
\begin{equation}
p_{\text{sp}}\left( y = (a, o) \middle| x; \theta \right) = \frac{\exp(f_v \cdot f_{t})}{\sum_{(a', o') \in {C}^s} \exp(f_v \cdot f_{t})},
\end{equation}

\noindent where $f_t$ is the text feature representation from the shifted soft prompt and \( \mathcal{C}^s \) denotes the set of seen compositions. To encourage the adapted soft prompt to generate text features that align with the target compositions, the cross-entropy over these probabilities is minimized to form the soft prompt alignment loss:

\vspace{-2.5mm}
\begin{equation}
\mathcal{L}_{\text{sp}} = - \frac{1}{| \mathcal{C}^s |} \sum_{(x, y) \in {C}^s} \log \left( p_{\text{sp}}\left( y = (a, o) \middle| x; \theta \right) \right).
\end{equation}

Next, we ensure that the fused features accurately reflect the retrieved prompts from the repository. The probability of this is defined as \( p_{\text{ret}} \) and apply the softmax function over \( {C}^s \):

\vspace{-4mm}
\begin{equation}
\textstyle
p_{\text{ret}}\bigl(y=(a,o)\mid x;\theta\bigr)
= \frac{\exp\bigl(f_{ret} \cdot f_{t \rightarrow v}\bigr)}%
       {\sum_{(a',o') \in {C}^s} 
        \exp\bigl(f_{ret} \cdot f_{t \rightarrow v}\bigr)}.
\end{equation}

\noindent The cross-entropy loss is then minimized over the class probabilities. The objective function is defined as:

\begin{equation}
\mathcal{L}_{\text{ret}} = - \frac{1}{| {C}^s |} \sum_{(x, y) \in {C}^s} \log \left( p_{\text{ret}}\left( y = (a, o) \middle| x; \theta \right) \right).
\end{equation}

The total loss function for training the model is then a weighted combination of the compositional, attribute, object, and soft prompt losses, where \( \lambda_{\text{att\_obj}} \) and \( \lambda_{\text{sp}} \) are hyperparameters that control the relative weight of the attribute-object loss and the soft prompt loss:

\vspace{-2mm}
\begin{equation}
\mathcal{L}_{\text{total}} = \mathcal{L}_{\text{ret}} + \lambda_{\text{att\_obj}} \left( \mathcal{L}_{\text{att}} + \mathcal{L}_{\text{obj}} \right) + \lambda_{\text{sp}} \mathcal{L}_{\text{sp}}.
\end{equation}

\subsection{Inference}
To predict the most likely attribute-object composition \(\hat{y}\) in a closed-world scenario, we select the label \((a, o)\) from the test set \({C}^{\text{test}}\) that maximizes the  probability \(p_{\text{sp}}\bigl(y = (a, o)\mid x;\theta\bigr)\):
\begin{equation}
\hat{y} \;=\; \underset{(a, o) \in {C}^{\text{test}}}{\arg\max} \; p_{\text{sp}}\bigl(y = (a, o)\mid x;\theta\bigr),
\end{equation}

\noindent where \(p_{\text{sp}}\) is computed following the same procedure in Eq.\,(8). Since the true attribute-object labels are unknown at inference, we construct text prompts for each \((a,o)\in {C}^{\text{test}}\) using our learned prefix, and select the pair yielding the highest \(p_{\text{sp}}\).

For open-world inference, ${C}$ expands to encompass all possible attribute-object pairs, making classification more challenging. As in prior works~\cite{nayak2022learning, bao2023prompting, lu2023decomposed}, a feasibility calibration step is applied by computing a similarity score \(p(a, o)\) for each candidate pair \((a, o)\). Any pair whose score falls below a threshold \(T\) is deemed infeasible and filtered out:
\begin{equation}
\hat{y} \;=\; \underset{(a, o)\,\in\,{C},\,p(a, o)\,\geq T}{\arg\max} \; p_{\text{sp}}\bigl(y = (a, o)\mid x;\theta\bigr). 
\end{equation}

This approach restricts the model to only consider attribute-object pairs deemed feasible. Selecting \(\hat{y}\) from the remaining pairs ensures that we capture the most probable composition for a given image, whether in a closed- or open-world setting.

\section{Experiments and Results}
\label{sec:experiment}
\subsection{Experimental Setup}

\noindent \textbf{Datasets.} We evaluate our model on three renowned CZSL datasets: MIT-States \cite{isola2015discovering}, UT-Zappos \cite{yu2014fine}, and C-GQA \cite{naeem2021learning}. MIT-States contains a variety of web-crawled images with 115 and 245 attributes and objects, respectively. 1262 seen compositions are used in training and 400 seen and unseen compositions used in testing. UT-Zappos is a smaller dataset containing images of 12 different types of shoes and 16 fine-grained attributes. C-GQA was built on top of the Stanford GQA dataset and contains a wide array of real life objects and attributes and possesses the most robust label space out of all three datasets, with over 800 seen and 900 unseen compositions in the test set.


\noindent \textbf{Metrics.} Following the setting of previous works \cite{li2024context, lu2023decomposed, xu2024gipcol}, we assess our model's performance using metrics tp focus on both seen and unseen compositions. Specifically, we evaluate accuracy for \textit{Seen (S)} and \textit{Unseen (U)} compositions under both closed-world and open-world scenarios, as these two cases offer insights into the model's generalization capabilities. Furthermore, we observe the \textit{Harmonic Mean (H)}. Given the inherent bias of zero-shot models toward seen compositions \cite{chao2016empirical, mancini2021open, min2020domain}, we analyze the trade-off between seen and unseen performance by plotting an accuracy curve across a bias range from $-\infty$ to $+\infty$. This allows us to compute the Area Under the Curve (AUC), the core metric reflecting the model’s overall capability.

\begin{table*}[t]
\centering
\begin{tabular}{l|cccc|cccc|cccc}
\toprule
\multirow{2}{*}{Method} & \multicolumn{4}{c|}{MIT-States} & \multicolumn{4}{c|}{UT-Zappos} & \multicolumn{4}{c}{C-GQA} \\
\cmidrule{2-13}
& S & U & H & AUC & S & U & H & AUC & S & U & H & AUC \\
\midrule
AoP \cite{nagarajan2018attributes} & 16.6 & 5.7 & 4.7 & 0.7 & 50.9 & 34.2 & 29.4 & 13.7 & - & - & - & - \\
LE+ \cite{naeem2021learning} & 14.2 & 2.5 & 2.7 & 0.3 & 60.4 & 36.5 & 30.5 & 16.3 & 19.2 & 0.7 & 1 & 0.1 \\
TMN \cite{purushwalkam2019task}  & 12.6 & 0.9 & 1.2	& 0.1 & 55.9 & 18.1	& 21.7& 8.4 & - & - & - & - \\
SymNet \cite{li2020symmetry}    & 21.4 & 7	& 5.8 & 0.8 & 53.3 & 44.6 & 34.5	& 18.5 & 26.7 & 2.2 & 3.3 & 0.4 \\
CompCos \cite{mancini2021open}   & 25.4	& 10.0 & 8.9 & 1.6	& 59.3	& 46.8 &	36.9 & 21.9 & - & - & - & - \\
CGE \cite{naeem2021learning} & 32.4 & 5.1	& 6.0 & 1.0 & 61.7 & 47.7 & 39.0 & 23.1 & 26.7 & 2.2 & 3.3 & 0.5\\
Co-CGE \cite{mancini2022learning} & 30.3 & 11.2	& 10.7 & 2.3 & 61.1 & 45.8 & 40.8 & 23.3 & 32.1 &	3.0 & 4.8 & 0.8\\
KG-SP \cite{karthik2022kg} & 28.4 & 7.5 & 7.4 & 1.3 & 61.8 & 52.1 &	42.3 & 26.5 & 31.5 & 2.9 & 4.7 & 0.8 \\
\midrule
CLIP \cite{radford2021learning}  & 30.1 & 14.3 & 12.8 & 3.0 & 15.7 & 20.6 & 11.6 & 2.2 & 7.5 & 4.6 & 4.0 & 0.3 \\
CSP \cite{nayak2022learning} & 46.3 & 15.7 & 17.4 &	5.7	& 64.1 & 44.1 & 38.9 & 22.7 & 28.7 & 5.2 & 6.9 & 1.2 \\
GIPCOL \cite{xu2024gipcol}  & \textbf{48.5} & 16.0 &17.9 &	6.3	& \textcolor{blue}{65.0} & 45.0 &	40.1 & 23.5 & 31.6 & 5.5 & 7.3 & 1.3\\
DFSP \cite{lu2023decomposed} & 47.5 & \textbf{18.5} & 19.3 & \textcolor{blue}{6.8} & \textbf{66.8} & \textcolor{blue}{60.0} & \textcolor{blue}{44.0} & \textcolor{blue}{30.3} & \textcolor{blue}{38.3} & \textcolor{blue}{7.2} & \textcolor{blue}{10.4} & \textcolor{blue}{2.4}\\
\midrule
VAPS (Ours) & \textcolor{blue}{48.3} & \textcolor{blue}{18.2} & \textbf{20.0} & \textbf{7.0} & 64.4 & \textbf{60.1} & \textbf{47.8} & \textbf{31.7} & \textbf{39.5} & \textbf{7.3} & \textbf{10.8} & \textbf{2.6} \\
\bottomrule
\end{tabular}
\caption{Open-World Results on MIT-States, UT-Zappos, and CGQA. The results are reported for Seen (S), Unseen (U), Harmonic Mean (H), and Area Under the Curve (AUC). Bold and blue indicate the first and second best results, respectively.}
\label{fig:open}
\end{table*}

\noindent\textbf{Implementation Details.} We utilize PyTorch 1.12.1 \cite{paszke2019pytorch} for the implementation of our model. The model is optimized using the Adam optimizer over the previously mentioned datasets. Both the image encoder and text encoder are based on the pretrained CLIP ViT-L/14 model, and the entire model is trained and evaluated on a single NVIDIA A100 GPU. We set $M = 20$ for UT-Zappos and $M = 30$ for MIT-States and C-GQA, as the latter two datasets contain a wider variety of attribute-object compositions. 

\subsection{Comparison with State-of-the-Arts}
Our method is compared to other state-of-the-art (SOA) CZSL methods, including: AoP \cite{nagarajan2018attributes}, LE+ \cite{naeem2021learning}, TMN \cite{purushwalkam2019task}, SymNet \cite{li2020symmetry}, CompCos \cite{mancini2021open}, CGE \cite{naeem2021learning}, Co-CGE \cite{mancini2022learning}, SCEN \cite{li2022siamese}, CLIP \cite{radford2021learning}, CSP \cite{nayak2022learning}, GIPCOL \cite{xu2024gipcol}, and DFSP \cite{lu2023decomposed}. The same data splits are used across each model and are based using CLIP's ViT-L/14 backbone. 

The main results for the Closed-World setting are reported in Table \ref{fig:closed}. We can observe that VAPS outperforms all other SOA methods on the UT-Zappos dataset, specifically with an increase of 2.6\% in classifying unseen compositions, a 5.0\% increase in harmonic mean, and a 3.9\% increase in AUC. When tested on the C-GQA dataset, our model demonstrates strength in seen accuracy with a 1.4\% improvement, harmonic mean with a 1.0\% increase and an increase in AUC to the previous SOA method by 0.5\%, further showcasing its robust performance across the benchmarks. Additionally, VAPS remains competitive on MIT-States, delivering best results in seen, unseen, and AUC. Table \ref{fig:open} showcases the results for the more challenging open-world scenario. An improved unseen accuracy, harmonic mean, and AUC can be observed UT-Zappos, while all metrics seen increases across CGQA. Once again, we show an increase in harmonic mean and AUC on the MIT-States dataset. We can attribute this success to the use of the visual prompt repository, which leverages learned visual semantics from the image encoder, as well as the prompt adapter, which shifts the soft prompt prefix for each individual image. Previous methods disregard enhancing the visual features from the image encoder while also assuming that one learned soft prompt prefix can generalize to all compositions. These outcomes emphasize the effectiveness of VAPS in both closed and open-world scenarios, where its visual prompt retrieval mechanism and soft prompt prefix adaptation results in consistent gains over all datasets against competing methods.

\subsection{Ablation Study}
To better understand the behavior of our model, we begin by conducting an ablation study to assess the contribution of each component branch. Additionally, we explore how varying the number of selected prompts from the visual repository affects performance. This study is performed on the UT-Zappos and MIT-States datasets. 

\vspace{1mm}
\noindent \textbf{Component Study.} We analyze how different branches interact in the proposed model in both closed-world and open-world settings in Table \ref{fig:ablation}. Specifically, we examine the cross-attention \textit{(ca)}, prompt adapter \textit{(pa)}, and prompt repository \textit{(pr)} components. We keep all configurations consistent throughout each experiment. When the \textit{pa} branch is excluded from the model, we observe a decrease across all metrics when compared to the full model. Over a 2\% decrease occurs on the seen and unseen composition accuracy and AUC drops 1.6\% on UT-Zappos in the open-world setting. For the closed-world setting, a decrease of 6.6\% and 3.7\% occur across unseen accuracy and AUC, respectively. Similarly, we see a decrease across all metrics on MIT-States when removing the \textit{pa} branch. Next, we perform an ablation to study the effect of the prompt repository by keeping only \textit{ca} and \textit{pa}. When removing \textit{pr}, all metrics decrease across the UT-Zappos dataset for both closed and open world scenarios. On MIT-States, a slight decrease of 0.4\% in AUC occurs; however, seen accuracy increases slightly by 0.1\%. Overall, we can see a higher variability in the results across the UT-Zappos dataset when ablating components compared to MIT-States. Retaining the prompt repository alongside the cross-attention and prompt adapter components ensures optimal performance, particularly in complex open-world scenarios. Furthermore, retaining all three branches of the model ensures that the core AUC metric remains at its highest level.

\begin{figure*}[t]
    \centering
    \includegraphics[width=\textwidth, height=0.25\textheight, keepaspectratio]{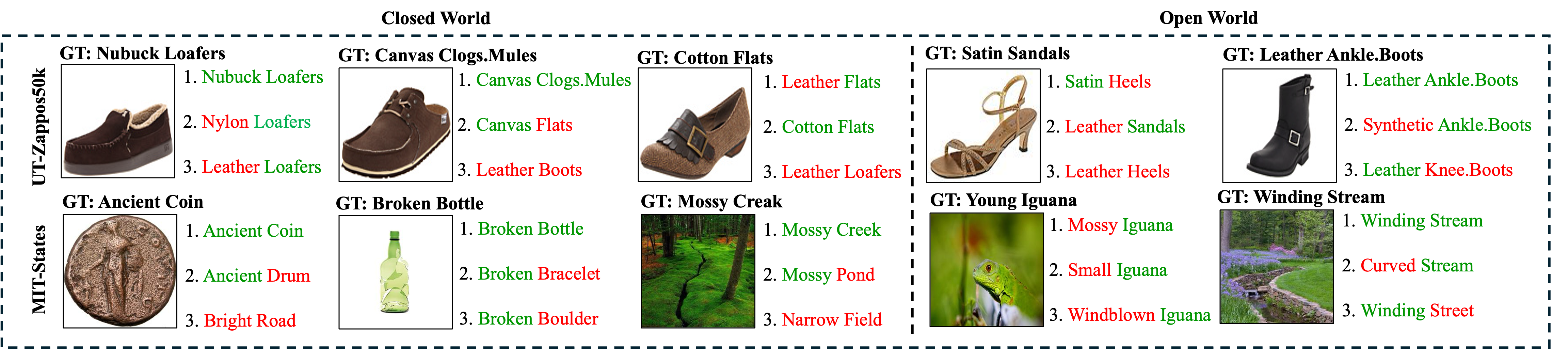} 
    \caption{\textit{Top-3 qualitative results. Each example displays the ground truth (GT) label and predictions, highlighting correct predictions in green and incorrect ones in red.}}
    \label{fig:QualFig}
\end{figure*}

\begin{table}[t!]
\centering
\resizebox{\columnwidth}{!}{
\begin{tabular}{lccc|cccc|cccc}
\toprule
 & \multicolumn{3}{c}{Components } & \multicolumn{4}{c}{MIT-States} & \multicolumn{4}{c}{UT-Zappos} \\
\cmidrule(lr){1-4} \cmidrule(lr){5-8} \cmidrule(lr){9-12}
& $pr$ & $ca$ & $pa$ & S & U & H & AUC & S & U & H & AUC \\
\midrule
\multirow{5}{*}{CW} & \checkmark & \checkmark & \checkmark & 48.4 & \textbf{52.2} & \textbf{38.2} & \textbf{21.2} & \textbf{64.5} & \textbf{74.3} & \textbf{53.8} & \textbf{40.1} \\
& \checkmark & \checkmark & & 47.9 & 51.5 & 37.3 & 20.5 & 62.2 & 72.1 & 52.2 & 38.5  \\
& & \checkmark  & \checkmark & \textbf{48.5} & 51.6 & 37.6 & 20.8 & 60.7 & 69.7 & 50.1 & 36.0 \\
\midrule
\multirow{5}{*}{OW} & \checkmark & \checkmark & \checkmark & 48.3 & \textbf{18.2} & \textbf{20.0} & \textbf{7.0} & \textbf{64.4} & \textbf{60.1} & \textbf{47.8} & \textbf{31.7} \\
& \checkmark & \checkmark & & \textbf{48.8} & 17.4 & 19.2 & 6.6 & 64.0 & 53.5 & 44.2 & 28.0 \\
&  & \checkmark & \checkmark & 48.7 & 17.5 & 19.5 & 6.8 & 61.8 & 53.2 & 43.0 & 26.9 \\
\bottomrule
\end{tabular}
}
\caption{Ablation of different model components and its effect on performance on Closed World (CW) and Open World (OW). Bold represents the best results.}
\label{fig:ablation}
\end{table}

\noindent \textbf{Repository Size and Number of Selected Prompts.} This study experiments with initiating the size of the repository and the number of selected prompts with different values. Through empirical analysis, it was observed the optimal visual repository size was between 20 and 30 prompts for each dataset. This range offers a sufficient diversity of learned prompts. Additionally, our analysis showed that selecting a single visual prompt for attributes and one for objects per image was optimal for compositional disentanglement. Our results show some variability between datasets. On MIT-States, increasing the repository size slightly improves harmonic mean and AUC, suggesting that a moderate diversity of prompts supports its range of compositions. For UT-Zappos, smaller repository sizes and selecting two prompts per image yield more noticeable AUC gains, likely due to one prompt representing attributes and one per image. This demonstrates that while both datasets benefit from the prompt repository, optimal configurations vary with dataset characteristics, showcasing the adaptability of our approach across different compositional settings.

\subsection{Qualitative Results}
To observe the robustness of our method, the top-3 qualitative results for selected images across UT-Zappos and MIT-States datasets are reported for both closed and open world settings, shown in Figure \ref{fig:QualFig}. For example, our model successfully classifies images of 'Nubuck Loafers' and 'Canvas Clogs' with the first prediction, but classifies 'Cotton Flats' correctly in the second prediction. However, the first prediction for 'Cotton Flats' was 'Leather Flats', which may be due to the brown color of the flats in the studied image. Similar results also occur in the MIT-States and UT-Zappos datasets. Our model correctly classifies all three of the images from MIT-States with its first prediction. For the 'Ancient Coin' image, despite the challenging visual similarities it may have with objects such as 'Ancient Drum', the model demonstrates strong attribute recognition by correctly identifying the 'Ancient' attribute, which helps it correctly classify the object. Similarly, for the 'Broken Bottle' image, although the texture and overall form might overlap with objects like 'Broken Boulder', the model effectively uses the 'Broken' attribute as a distinguishing feature, allowing it to make informed predictions. 

Our model's robustness is further demonstrated in the open world setting. For instance, it successfully differentiates between visually similar but distinct material, such as 'Leather' and 'Synthetic', when classifying 'Ankle Boots'. Similarly, the model was able to classify the image of the 'Winding Stream' with high accuracy. However, some incorrect first predictions are seen in each dataset, such as predicting the image of the 'Iguana' as 'Mossy' instead of 'Young'. Also, the prediction of a 'Windblown Iguana' highlights the challenge posed by the open-world scenario since this attribute is not plausible for the given object. Overall, these qualitative results represent the effectiveness of our method in both closed and open world settings. By focusing on key object features and attribute details, our model is able to make accurate classifications, even when confronted with unseen compositions or visually complex scenarios. 

\begin{table}[t!]
\centering
\resizebox{\columnwidth}{!}{
\begin{tabular}{l|l|cccc|cccc}
\toprule
Rep. Size & Selected N & \multicolumn{4}{c|}{MIT-States} & \multicolumn{4}{c}{UT-Zappos} \\
\cmidrule(lr){3-6} \cmidrule(lr){7-10}
& & S & U & H & AUC & S & U & H & AUC \\
\midrule
20 & 2 & 49.2 & 50.5 & 37.2 & 20.5 & 63.5 & 72.7 & 53.8 & 40.1\\
20 & 4 & 49.3 & 50.4 & 37.1 & 20.6 & 64.4 & 70.1 & 47.1 & 35.1 \\
30 & 2 & 48.4 & 52.2 & 38.2 & 21.2 & 62.4 & 72.3 & 49.9 & 36.9 \\
30 & 4 & 48.5 & 52.1 & 38.2 & 20.8 & 61.0 & 72.6 & 49.3 & 35.8\\
\bottomrule
\end{tabular}
}
\caption{Comparison of datasets with varying repository size and number of selected prompts in closed-world setting.}
\label{fig:num_prompts}
\end{table}

\section{Conclusion}
\label{sec:conclusion}
In this paper, we propose a Visual Adaptive Prompting System (VAPS) to bridge the gap between semantic and visual features for CZSL. By leveraging a dynamic visual prompt repository and a similarity-based retrieval mechanism to select relevant visual prompts based on attribute and objects, VAPS enhances the model's ability to generalize to unseen compositions. Furthermore, we propose dynamically adapting the learnable prompt prefix based on the image features derived from CLIP's image encoder. Our experiments on benchmark datasets demonstrate that VAPS achieves state-of-the-art performance on metrics across the UT-Zappos and CQGA datasets, highlighting its capability to disentangle and recompose visual features effectively. 





{
    \small
    \bibliographystyle{ieeenat_fullname}
    \bibliography{main}
}

\end{document}